\documentclass[sigconf]{acmart}

\usepackage{graphicx}
\usepackage{balance}
\usepackage{algorithm} 
\usepackage{algorithmic}
\usepackage{hyperref}
\usepackage[normalem]{ulem}
\usepackage{float}
\useunder{\uline}{\ul}{}
\usepackage{stfloats} 
\usepackage{tcolorbox}
\tcbuselibrary{most}

\copyrightyear{2025}
\acmYear{2025}
\acmDOI{XXXXXXX.XXXXXXX}
\acmConference[KDD 2025]{Proceedings of the 2025 ACM Conference on Knowledge Discovery and Data Mining}{August 2025}{Toronto, Canada}

\newcommand{\surveyx}{\textsc{SurveyX}}
\definecolor{rred}{HTML}{941100}
\definecolor{bblue}{HTML}{0000ff}
\newtcbox{\cb}[1][{red!75!black}]{
    on line,
    colback=#1,
    coltext=white,
    boxrule=0.4pt,        
    left=0.8pt,           
    right=0.8pt,
    top=0.1pt,            
    bottom=1pt,
    arc=0pt,              
    before upper={\rule[-0.8pt]{0pt}{9pt}}, 
    fontupper=\sffamily\small, 
    boxsep=0pt,           
    nobeforeafter         
}

\begin{document}

\title{\surveyx{}: Academic Survey Automation via Large Language Models}

\author{Xun Liang}
\email{xliang@ruc.edu.cn}
\authornote{These authors contributed equally to this research.}
\affiliation{%
  \institution{Renmin University of China}
  \city{Beijing}
  \country{China}
}
\author{Jiawei Yang}
\authornotemark[1]
\email{j1aweiyang@ruc.edu.cn}
\affiliation{%
  \institution{Renmin University of China}
  \city{Beijing}
  \country{China}
}

\author{Yezhaohui Wang}
\email{yezhaohuiwang@gmail.com}
\authornotemark[1]
\affiliation{%
  \institution{Northeastern University}
  \city{Shenyang}
  \country{China}
}

\author{Chen Tang}
\email{tangc@iaar.ac.cn}
\authornotemark[1]
\affiliation{%
  \institution{Institute for Advanced Algorithms Research}
  \city{Shanghai}
  \country{China}
}
\author{Zifan Zheng}
\email{zzhe0348@uni.sydney.edu.au}
\affiliation{%
  \institution{The University of Sydney}
  \city{Sydney}
  \country{Australia}
}
\author{Shichao Song}
\email{songshichao@ruc.edu.cn}
\affiliation{%
  \institution{Renmin University of China}
  \city{Beijing}
  \country{China}
}
\author{Zehao Lin}
\email{linzh@iaar.ac.cn}
\affiliation{%
  \institution{Institute for Advanced Algorithms Research}
  \city{Shanghai}
  \country{China}
}
\author{Yebin Yang}
\email{yangyb@iaar.ac.cn}
\affiliation{%
  \institution{Institute for Advanced Algorithms Research}
  \city{Shanghai}
  \country{China}
}
\author{Simin Niu}
\email{niusimin@ruc.edu.cn}
\affiliation{%
  \institution{Renmin University of China}
  \city{Beijing}
  \country{China}
}
\author{Hanyu Wang}
\email{hy.wang@ruc.edu.cn}
\affiliation{%
  \institution{Renmin University of China}
  \city{Beijing}
  \country{China}
}
\author{Bo Tang}
\email{tangb@iaar.ac.cn}
\affiliation{%
  \institution{Institute for Advanced Algorithms Research}
  \city{Shanghai}
  \country{China}
}
\author{Feiyu Xiong}
\email{xiongfy@iaar.ac.cn}
\affiliation{%
  \institution{Institute for Advanced Algorithms Research}
  \city{Shanghai}
  \country{China}
}
\author{Keming Mao}
\email{maokm@mail.neu.edu.cn}
\affiliation{%
  \institution{Northeastern University}
  \city{Shenyang}
  \country{China}
}
\author{Zhiyu Li}
\email{lizy@iaar.ac.cn}
\authornote{Corresponding author: lizy@iaar.ac.cn.}
\affiliation{%
  \institution{Institute for Advanced Algorithms Research}
  \city{Shanghai}
  \country{China}
}

\renewcommand{\shortauthors}{Xun Liang et al.}
\begin{abstract}
Large Language Models (LLMs) have demonstrated exceptional comprehension capabilities and a vast knowledge base, suggesting that LLMs can serve as efficient tools for automated survey generation. However, recent research related to automated survey generation remains constrained by some critical limitations like finite context window, lack of in-depth content discussion, and absence of systematic evaluation frameworks. Inspired by human writing processes, we propose \surveyx{}, an efficient and organized system for automated survey generation that decomposes the survey composing process into two phases: the Preparation and Generation phases. By innovatively introducing online reference retrieval, a pre-processing method called AttributeTree, and a re-polishing process, \surveyx{} significantly enhances the efficacy of survey composition. Experimental evaluation results show that \surveyx{} outperforms existing automated survey generation systems in content quality (0.259 improvement) and citation quality (1.76 enhancement), approaching human expert performance across multiple evaluation dimensions. Examples of surveys generated by \surveyx{} are available on \href{www.surveyx.cn}{our project website}\footnote{\url{http://www.surveyx.cn}}.
\end{abstract}



\keywords{Automated Survey Generation, Literature Synthesis, Large Language Models, NLP}

\received{10 February 2025}

\maketitle

\begin{figure}[h]
  \centering
  \includegraphics[width=\columnwidth]{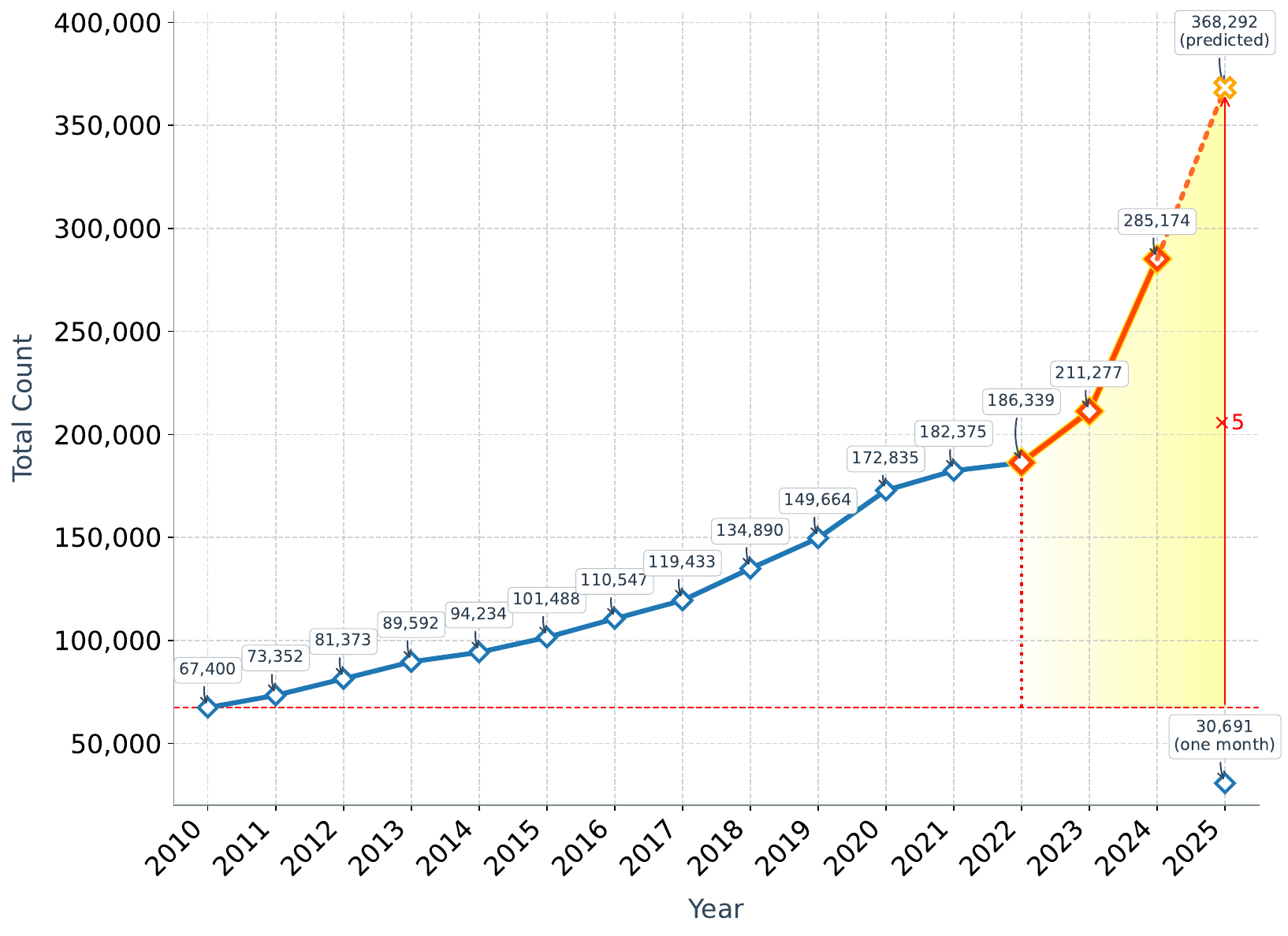}
  \caption{The number of papers received annually by the arXiv website from 2010 to 2025, with data sourced from our arXiv database. The projected number of submissions for 2025 is anticipated to be five times greater than that of 2010.}
  \label{fig:paper_trends}
\end{figure}
\section{Introduction} \label{sec:introduction}
In recent years, computer science has advanced rapidly across various fields \cite{bick2024rapid,gupta2023recent,loakman2024train}. Statistics indicate that arXiv.org\footnote{\url{https://arxiv.org/}} receives about a thousand new papers daily. Figure \ref{fig:paper_trends} highlights a remarkable trend: over the past three years (2022-2024), the number of papers published on arXiv has increased substantially, rising from 186,339 to 285,174, representing a growth of over 50\%. It is projected that this number will further climb to approximately 368,292 in 2025. However, the exponential growth of literature has made it increasingly challenging for researchers attempting to comprehend the technological evolution and developmental trajectories of specific subfields from the ground up \cite{zhong2024solution}. Surveys are instrumental in elucidating the current state of research and the historical progression within a given topic \cite{10.1145/3641289,zhao2023survey,ZHENG2025101176}. However, the workload involved in manually writing these surveys is continually increasing, threatening the ability to maintain comprehensive coverage and high quality of these surveys. Against the backdrop of information overload, there is an urgent need to develop efficient systems for automated survey generation. The rise of Large Language Models (LLMs) has rendered automated survey generation a viable approach \cite{hurst2024gpt,team2023gemini,kevian2024capabilities}. Trained on large-scale textual corpora, LLMs possess the capability to produce text that is both fluent and logically coherent \cite{guo2023evaluating,minaee2024large,NEURIPS2023_43e9d647,tang2024cross}. Despite this potential, leveraging LLMs for automated survey generation poses several challenges, which can be categorized into two levels:

\textbf{1) Technical Challenges:} 

$\bullet$ LLMs primarily rely on their internal knowledge bases for text generation. However, effective survey composition requires comprehensive and accurate support from up-to-date references, whereas knowledge stored within LLMs may become outdated and occasionally provide incorrect reference information \cite{onoe-etal-2022-entity,li-etal-2023-large,zhang2023siren,huang2023survey}. This limitation causes negative effects on the academic rigor and credibility of the generated surveys. Consequently, relying solely on LLMs makes it difficult to generate high-quality surveys.

$\bullet$ Current mainstream LLMs encounter limitations due to their context window sizes \cite{li2024long,li2023how,kaddour2023challenges}. For instance, GPT-4o has a context window of 128K tokens, while Claude 3.5 has 200K tokens. Crafting a comprehensive survey typically entails citing hundreds of references, each averaging around 10K tokens. This scale far exceeds the context window capacity of existing LLMs, posing a challenge to directly equipping them with all necessary references for generating high-quality surveys.

\textbf{2) Application Challenges:}

$\bullet$ The creation of surveys relies on a substantial number of references, necessitating timely content retrieval via online methods. However, there is currently a lack of effective tools that can efficiently procure large volumes of the latest and highly pertinent references, thus limiting the broader application of automated survey generation.

$\bullet$ Unlike traditional Natural Language Generation (NLG) tasks, the evaluation of surveys lacks unified metrics and standardized benchmarks \cite{AutoSurvey_24_NIPS_Westlake}. The absence of evaluation frameworks hampers effective quality assessment of automatically generated surveys, thereby constraining their applicability in large-scale academic settings.

Existing similar works have provided some solutions to the aforementioned challenges, yet notable deficiencies persist, particularly in the following areas:

\begin{itemize}
\item Exisiting retrieval methodologies contain inherent limitations. In existing works, \citet{AutoSurvey_24_NIPS_Westlake} only supports offline retrieval and cannot access the latest references, leading to a lack of timeliness.

\item The methods for pre-processing references are often inadequate. In existing works, \citet{AutoSurvey_24_NIPS_Westlake} only utilizes partial information from the references (titles and abstracts), overlooking a significant amount of crucial content.

\item The generated surveys lack diversity in their expression. Existing methodologies are restricted to generating text-based surveys and fall short in including visual elements such as figures and tables, diminishing the readability of the results.
\end{itemize}

To address these deficiencies, we propose \surveyx{}, an efficient and well organized system for automated survey generation. \surveyx{}  divides the composing process of surveys into two phases: the Preparation Phase and the Generation Phase. In the Preparation Phase, \surveyx{} employs retrieval algorithms to search and filter highly relevant references from the internet based on the given survey topic. It employs a reference pre-processing method, termed AttributeTree, to distill key information from the references, constructing a reference materials database for efficient retrieval through the Retrieval Augmented Generation (RAG) technique. In the Generation Phase, \surveyx{} utilizes the information obtained in the previous phase to sequentially generate the outline and main body of the survey, ensuring that the generated survey has a clear structure and accurate content. Additionally, tables and figures are incorporated to enrich the survey's presentation. Beyond the generation process, \surveyx{} extends the evaluation framework proposed by \citet{AutoSurvey_24_NIPS_Westlake} by incorporating additional evaluation metrics, thus aiding subsequent related research.

To sum up, our main contributions are as follows:
\begin{itemize}

\item We propose an efficient reference retrieval algorithm capable of expanding keywords based on a given topic, substantially broadening the retrieval scope. Additionally, a 2-step filtering method is employed to eliminate papers with low relevance, leaving only high-quality references that comprehensively cover the topic.
\item We design a reference pre-processing method called AttributeTree, which efficiently extracts key information from documents. This method significantly enhances the information density of reference materials, improving LLMs’ comprehension and optimizing their context window usage.
\item We introduce an outline generation method named Outline Optimization, which generates outlines based on hints and employs a ``separate-then-reorganize'' step to eliminate redundancy. This method results in outlines with more rigorous logic and clearer structure.
\item We expand the expressive format of the generated survey to include figures and tables alongside text,  enriching the presentation and improving readability.
\item We augment the evaluation framework by introducing additional metrics to assess the quality of the generated surveys and retrieved references. The evaluation results demonstrate that the surveys generated by \surveyx{} outperform existing works across multiple metrics, closely aligning with human expert performance.
\end{itemize}

\begin{figure*}[ht]
  \centering
  \includegraphics[width=\textwidth]{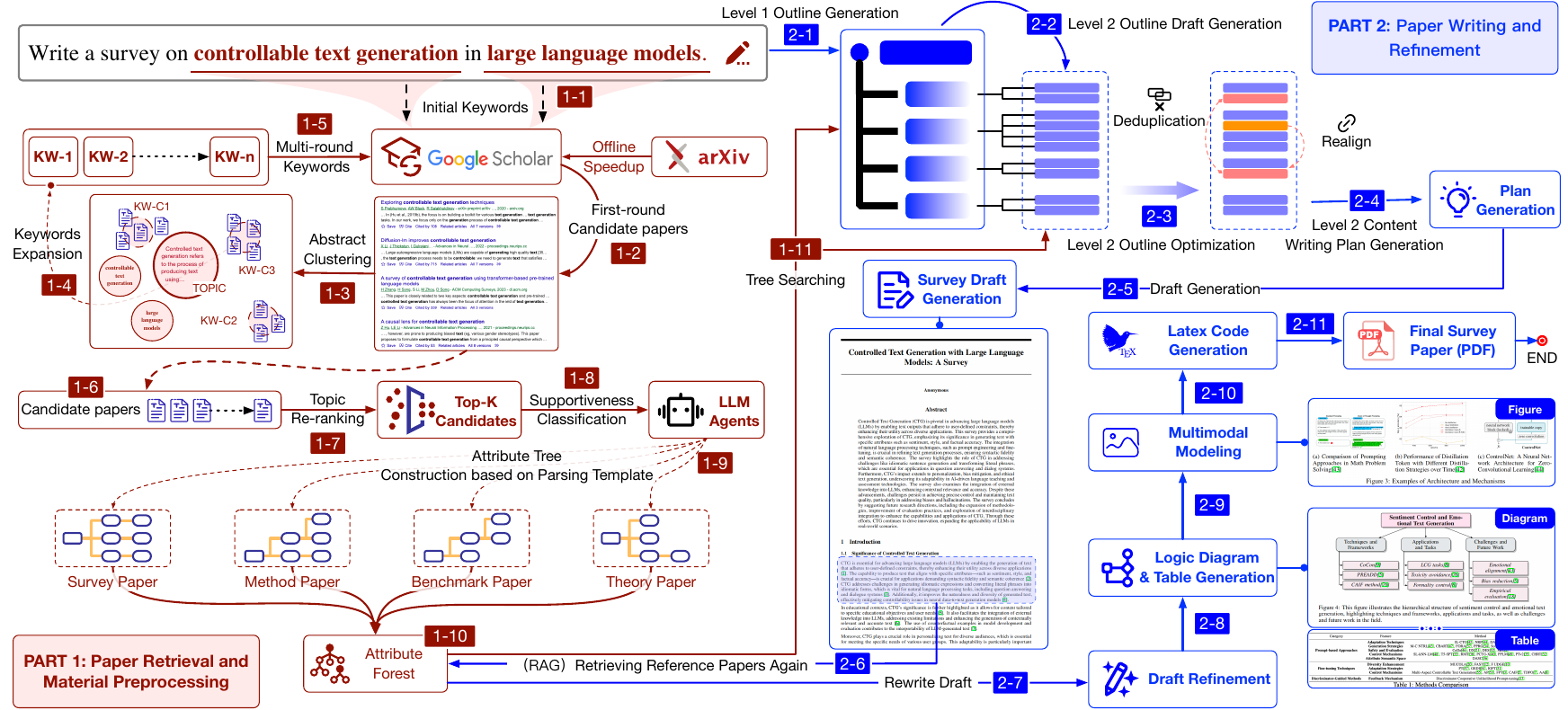}
  \caption{Pipeline of \surveyx{}.}
  \label{fig:frame}
\end{figure*}
\section{Related Work} 

\textbf{Long-form Text Generation.}
Although LLMs excel in traditional NLG tasks, generating long-form, well-structured, coherent, and logically organized text using LLMs still remains a persistent challenge \cite{tan2024proxyqa,kumar2024longlamp,wu2024spinning,min-etal-2023-factscore,que2024hellobench,dong-etal-2024-bamboo,tang-etal-2023-enhancing}. To address this problem, some studies have sought to employ planning strategies. For instance, \citet{tan-etal-2021-progressive} proposed a method that first produces domain-specific content keywords, and then progressively refines them into complete passages in multiple stages. Similarly, \citet{liang2024integrating} used a series of auxiliary training tasks to endow LLMs with the skills to plan and structure long-form documents before generating the final full article. Some other researches focus on the generation of long-form text in a specific format, such as Wikipedia articles, surveys, commentaries, etc \cite{zhang2024retrieval, wu2024automated, tang-etal-2022-etrica,tang-etal-2023-improving}. \citet{shao-etal-2024-assisting} developed STORM, a system that models the pre-writing stage by discovering diverse perspectives, simulating expert conversations, and curating information to create an outline, which is then used to generate a complete Wikipedia article. \citet{AutoSurvey_24_NIPS_Westlake} introduced AotuSurvey, a framework for generating surveys consisting of initial retrieval, outline generation, parallel subsection drafting, integration, and rigorous evaluation. In comparison, our work addresses existing shortcomings by enhancing the retrieval scenarios, optimizing the reference pre-processing methods, expanding the expressive forms of the generated surveys, and improving overall quality.

\textbf{Retrieval Augmented Generation (RAG).}
RAG technique is used to help LLMs access external knowledge, thereby enhancing their text generation capabilities \cite{gao2023retrieval, fan2024survey, huang2024survey, hu2024rag}. This technique has proven especially useful in tasks that require up-to-date or domain-specific information, such as QA (Question Answering) \cite{ke-etal-2024-bridging,asai2024selfrag}, Dialog Generation \cite{rajput2024recommender,linra}, Reasoning \cite{cheng-etal-2023-uprise,zhang-etal-2023-iag}, etc. Beyond traditional NLG tasks, RAG is also used for long-form text generation, as this task often requires handling extensive external knowledge \cite{AutoSurvey_24_NIPS_Westlake,shao-etal-2024-assisting,zhang-etal-2023-iag}. Compared to previous work that directly used the raw text of references as the retrieval data source, our work significantly improves retrieval efficiency and context window utilization by converting it into attribute trees.

\section{Methodology}
In this section, we delineate the pipeline of \surveyx{}, which includes two main phases: Preparation Phase (section \ref{subsec:Preparation Phase}) and Generation Phase (section \ref{subsec:Generation Phase}). The primary task of the Preparation Phase is to collect the related references needed for composing the survey and pre-processing them to construct a reference materials database for retrieval using RAG technique. The Generation Phase undertakes the creation of both the outline and main body of the survey based on these materials. This phase also involves the post-processing refinement of the initial draft to enhance its overall quality and augment its expressive presentation. An overview of this system is depicted in \autoref{fig:frame}.

\subsection{Preparation Phase} \label{subsec:Preparation Phase}
Preparation Phase consists of two stages: the references acquisition stage and the references pre-processing stage.

\subsubsection{References Acquisition}
Considering the current shortage of effective tools for efficiently acquiring a large volume of highly relevant references, we independently develop a module for acquiring references. This module comprises two separate components: \textbf{retrieval data source} and \textbf{retrieval algorithms}. 

The \textbf{retrieval data source} includes both offline and online data sources. The offline data source consists of 2,632,189\footnote{Data as of February 10, 2025.} available papers downloaded from the arXiv.org, with new papers added daily. The online data source is a self-developed crawler system based on Google Scholar. Using the offline data source facilitates quick access to existing references, thereby saving time, while the online data source enables the acquisition of the latest and multi-source references. The combination of both ensures efficiency and timeliness. 

The \textbf{retrieval algorithm} is divided into two steps: reference recall and reference filter (steps \cb[rred]{1-1}\textasciitilde
\cb[rred]{1-6} and steps \cb[rred]{1-7}\textasciitilde\cb[rred]{1-8} in Figure \ref{fig:frame}). (1) The goal of the reference recall step is to avoid missing any references related to the topic as much as possible. (2) The paper filter step aims to filter out references unrelated to the topic as much as possible. To achieve (1), we design a method called the Keyword Expansion Algorithm. The pseudo code can be found at Algorithm \ref{keyword_Expansion_algo}. The algorithm initializes the keyword pool with an initial keyword $K_0$. After searching references by newly added keywords, it performs semantic clustering on the abstracts of all retrieved references $Doc$ and summarizes keywords for each cluster $C_i$. The obtained keywords $K_{c}$ are then semantically compared with the existing keyword pool $K_{pool}$ and the topic $T$, the most appropriate keyword $k^*$ based on semantic distance is added to the keyword pool. This process is repeated until the number of $Doc$ reaches the threshold $\theta$ (set as 1000).

\begin{algorithm}
\caption{Keyword Expansion Algorithm}
\begin{algorithmic}[1]
\small
\STATE \textbf{Input:} initial keywords set $K_0$, topic $T$
\STATE \textbf{Output:} retrieved documents $Doc$
\STATE keyword pool: $K_{\text{pool}} \leftarrow K_0$
\WHILE{LengthOf($Doc$) $< \theta$}
    \STATE $k \leftarrow$ newly added keywords in $K_{\text{pool}}$
    \STATE $Doc \leftarrow Doc + \text{Search}(k)$
    \STATE $n \leftarrow \text{LengthOf}(K_{\text{pool}})$
    \STATE $C \leftarrow \text{Cluster}(Doc, n+1)$
    \STATE $K_{c} \leftarrow \text{ExtractKeywords}(C)$
    \STATE $Dist \leftarrow \text{CalcDist}(K_{c}, K_{\text{pool}}, T, \text{Weights})$
    \STATE $k^* \leftarrow \text{Select}(Dist, K_{\text{pool}})$ 
    \STATE $K_{\text{pool}} \leftarrow K_{\text{pool}} + k^*$
\ENDWHILE
\STATE \textbf{Return:} $Doc$
\end{algorithmic}
\label{keyword_Expansion_algo}
\end{algorithm}

\textit{Weights} are the semantic distance weights. Considering that all keywords should more closely revolve around the topic, we double the semantic distance weight of the topic. The \textit{Select} function, when choosing candidate words, assumes that the best candidate word should have the smallest possible average distance to the existing keywords while also having the maximum furthest distance. Specifically:

\begin{equation}
k^* = \arg\min_{k_c \in K_C} \big( R_1(k_c) + R_2(k_c) \big)
\label{eq:k_star}
\end{equation}

$R_1$ and $R_2$ represent different ranking calculation methods. Their calculation methods are as follows:
\begin{equation}
\resizebox{0.4\textwidth}{!}{
$R_1(k_c) = \text{rank}_{K_c}\left( \frac{1}{|K_{pool}|} \sum\limits_{k_e \in K_{pool}} \operatorname{cos\_sim}(E(k_c), E(k_e)) \right)$
}
\hfill
\end{equation}

\begin{equation}
R_2(k_c) = \text{rank}_{K_c} \left( -\max_{k_e \in K_{pool}} \operatorname{cos\_sim} (E(k_c), E(k_e)) \right)
\end{equation}
Here, $\text{rank}(\cdot)$ refers to the Rank Function, and $E(\cdot)$ represents the embedding model.
To achieve (2), we propose a 2-step filtration algorithm. The first step uses an embedding model to calculate the semantic relevance between the topic and abstracts of references, selecting the Top-K references most relevant to the topic, which serves as coarse-grained filtration. The second stage uses LLMs for more precise semantic filtering, serving as fine-grained filtration. This algorithm ensures that the selected papers maintain a high degree of relevance to the research topic, thereby enhancing the quality of the survey. 
\subsubsection{References Pre-processing}
After obtaining the references, they need to be preprocessed for use in subsequent stages. A naive and common pre-processing method is to directly provide the full text of the references to LLMs for generating the surveys. However, we believe this approach suffers from low context window utilization and inefficient extraction of key information. We have noticed that before composing a survey, people often categorize the necessary references and organize all potentially useful information. Based on this observation, we design a reference pre-processing method called AttributeTree. Specifically, we designed different attribute tree templates for different types of references in advance. Using these templates, we can efficiently extract key information from the references. The attribute trees of all references are combined into an attribute forest, which represents all the reference materials required for composing a survey (steps \cb[rred]{1-9}\textasciitilde\cb[rred]{1-10} in Figure \ref{fig:frame}).
Examples of attribute tree templates can be found in the Appendix \ref{sec:Attribute Tree Templates}. This method significantly increases the information density of the reference materials and efficiently utilizes the context window of LLMs, thus laying a solid foundation for composing high-quality surveys.

\subsection{Generation Phase}
\label{subsec:Generation Phase}
After obtaining all the reference materials needed for writing the survey, the generation phase begins. This phase is divided into three stages: \textbf{outline generation}, \textbf{main body generation}, and \textbf{post refinement}.
\begin{figure}[h]
  \centering
  \includegraphics[width=\columnwidth]{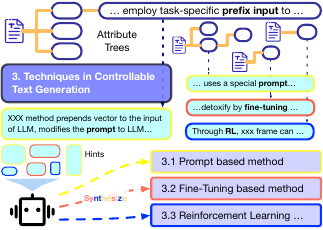}
  \caption{An example of generating secondary outlines. LLMs first generate hints based on the attribute tree to guide the generating of the secondary outline. Then, by synthesizing all hints, LLMs identify the most suitable entry points to determine the segmentation strategy and generate the secondary outline.}
  \label{fig:sub_frame}
\end{figure}
\subsubsection{Outline Generation}
Outline generation is the most crucial stage in the entire survey generation process. A well-structured, logically organized, and focused outline ensures a comprehensive, cohesive, and informative survey. Our preliminary experiments indicate that, when generating the primary outline, utilizing only the internal knowledge of LLMs is sufficient to produce a human-level primary outline. In contrast, generating the secondary outline is more challenging, and relying solely on the internal knowledge of LLMs is inadequate. We observed that humans typically categorize or summarize references to guide the writing of the survey. Inspired by this, we designed an outline generation method named Outline Optimization (steps \cb[bblue]{2-1}\textasciitilde\cb[bblue]{2-3} in Figure \ref{fig:frame}). This method consists of two steps. 

For the first step, this method initially lets LLMs generate hints corresponding to the secondary outline based on the attribute tree of each reference. A hint can be considered as a form of guiding content that assists LLMs in better understanding the key information a particular reference can provide in constructing a systematic framework\footnote{``systematic framework'' refers to the framework proposed in a survey for organizing and synthesizing all the references.}. Then, LLMs sequentially generate the secondary outline based on these hints. An example of this step is shown in Figure \ref{fig:sub_frame}. Through this step, LLMs can more accurately identify the commonalities and differences among various references and more efficiently synthesize and organize these references, thereby improving the generation of secondary outlines. 

For the second step, this method begins by separating all secondary outlines from the primary outlines, retaining only their headings to facilitate deduplication by the LLMs. Then, use LLMs to reorganize the deduplicated secondary outlines into the primary outline. Through this step, the method effectively addresses the redundancy present in the generated secondary outlines.

With the help of the outline optimization method, we can generate survey outlines with logical rigor and clear structure.

\subsubsection{Content Generation}
During the content generation stage, inspired by the human writing process for surveys, we continue to use the hint-based method employed during the outline generation stage to enhance the depth of the generated content (steps \cb[bblue]{2-4}\textasciitilde\cb[bblue]{2-5} in Figure \ref{fig:frame}). Unlike before, the hints used here are derived from the secondary outline, aiming to guide the generation of the main body content. LLMs will generate the main body content on a subsection\footnote{One secondary outline corresponds to one subsection.} basis using these hints and the outline produced in the previous stage. Meanwhile, when the LLM is writing a specific subsection, it can view the content of other subsections. This ensures that the content currently being generated remains consistent with the content already generated to some extent.

\subsubsection{Post Refinement}
In the human process of writing surveys, revising and polishing the draft to enhance its quality is common. Based on this, we design a post refinement stage, which polishes the initial draft with two main goals: (1) to improve the quality of the content, including citation quality, textual fluency, and consistency of expression; (2) to add figures and tables to enrich the presentation of the survey. These two objectives are achieved by an RAG-based rewriting module and a graph and table generation module, respectively.

\textbf{RAG-based Rewriting Module.}  Even though strategies were used in the content generation stage to ensure content consistency in the generated survey, these efforts are not completely sufficient. Additionally, the accuracy of citations in the generated survey has not been verified. To address these issues, we design a RAG-based rewriting module to revise the survey content. This module first uses the paragraphs from the initial draft as queries to retrieve reference materials from the attribute forest. Subsequently, it constructs a prompt based on these materials to rewrite the paragraph using LLMs (steps \cb[bblue]{2-6}\textasciitilde\cb[bblue]{2-7} in Figure \ref{fig:frame}). The rewriting process involves two aspects: first, removing irrelevant citations and adding highly relevant ones to the paragraph, and second, polishing the paragraph by considering its context.
This module not only significantly enhances the citation quality but also ensures the content consistency of the generated survey.

\textbf{Figure and Table Generation Module.}
 Surveys that consist solely of text often lack expression diversity, which can limit their effectiveness in conveying information. To address this issue, we designed a figure and table generation module (steps \cb[bblue]{2-8}\textasciitilde\cb[bblue]{2-11} in Figure \ref{fig:frame}). Inspired by Napkin\footnote{\url{https://app.napkin.ai/}}, We construct several information extraction templates, each corresponding to a specific figure or table generation template (the generation templates are scripts). Based on the information extraction templates, we use LLMs to extract the necessary information from the attribute tree of the references for generating figures or tables. Subsequently, the generation templates automatically construct the corresponding figures or tables based on the extracted information. Additionally, on a subchapter basis, we leverage Multimodal Large Language Models (MLLMs) combined with context to retrieve figures from references. If a figure effectively supports the content of a subsection, we incorporate it to enhance clarity and expressiveness.

 Through the aforementioned modules, we have constructed a comprehensive and systematic post refinement stage that significantly enhances the overall quality of the generated survey while expanding its expressive format.

\begin{table*}[hb]
\caption{Content quality evaluation results of naive RAG, Autosurvey, SurveyX and Human writing. All LLM-Agent is GPT-4o.}

\begin{tabular}{cccccccc|ccc}
\hline
Model            &  & Coverage      & Structure     & Relevance     & Synthesis     & \begin{tabular}[c]{@{}c@{}}Critical\\ Analysis\end{tabular} & Avg            & \multicolumn{1}{l}{Recall} & \multicolumn{1}{l}{Precision} & \multicolumn{1}{l}{F1} \\ \hline
naive RAG        &  & 4.40          & 3.66          & 4.66          & 3.82          & 2.82                                                        & 3.872          & 68.79                      & 61.97                         & 65.20                        \\
AutoSurvey       &  & 4.73          & 4.33          & 4.86          & 4.00          & 3.73                                                        & 4.331          & 82.25                      & 77.41                         & 79.76                        \\
\textbf{\surveyx{}} &  & \textbf{4.95} & \textbf{4.91} & \textbf{4.94} & \textbf{4.10} & \textbf{4.05}                                               & \textbf{4.590} & \textbf{85.23}             & \textbf{78.12}                & \textbf{81.52}               \\ \hline
Human            &  & 5.00          & 4.95          & 5.00          & 4.44          & 4.38                                                        & 4.754          & 86.33                      & 77.78                         & 81.83                        \\ \hline
\end{tabular}
\label{tab:content}

\end{table*}

\section{Experiments}
\subsection{Evaluation Metrics}

\paragraph{Automatic evaluation} In terms of content quality evaluation, we expanded the evaluation dimensions presented in \cite{AutoSurvey_24_NIPS_Westlake} by incorporating synthesis and critical analysis into the evaluation metrics. Synthesis evaluates the ability to interconnect disparate studies, identify overarching patterns or contradictions, and construct a cohesive intellectual framework beyond individual summaries, while critical analysis examines the depth of critique applied to existing studies, including the identification of methodological limitations, theoretical inconsistencies, and research gaps.

For the evaluation of citation quality, we adopted the Citation Recall and Citation Precision metrics proposed by \cite{AutoSurvey_24_NIPS_Westlake}. The former evaluates whether each statement in the generated text is fully supported by the cited references, while the latter evaluates for the presence of irrelevant citations. In addition, we introduced the F1 Score metric to provide a more comprehensive evaluation of citation quality.

Regarding the evaluation of reference relevance, we identify an existing gap in the literature, as this aspect has not been addressed in previous studies. To fill this void, we propose a novel set of metrics specifically designed to assess the relevance of retrieved references. The details of this proposed metric set are as follows:

(1) IoU (Insertion over Union) This metric measures the similarity between machine-retrieved and human-retrieved references by calculating the degree of overlap between the two. The calculation method is as follows
$$
IoU = \frac{Doc_{human} \cap Doc_{machine}}{Doc_{human} \cup Doc_{machine}}
$$

(2) \text{Relevance}\textsubscript{\textit{semantic}}
(semantic-based reference relevance). This metric is based on an embedding model, calculating the cosine similarity between the embeddings of the retrieved references and the topic to measure their relevance. The calculation method is as follows:
$$
Relevance_{semantic} = \frac{1}{|Doc|}\sum_{d \in Doc}{{\operatorname{cos\_sim}}(E(d), E(topic))}  
$$

(3) \text{Relevance}\textsubscript{\textit{LLM}}
(LLM-based reference relevance).
This metric is designed by constructing prompts to leverage LLMs to directly evaluate the relevance of the retrieved references to the topic. The calculation method is as follows:
$$
Relevance_{LLM} = \frac{1}{|Doc|}\sum_{d \in Doc}{\mathbb{I}_{\text{relevant}}(LLM(Prompt(d,topic)))}
$$

The prompts used in the evaluation are presented in the Appendix. \ref{subsec:Evaluation Prompt}
\paragraph{Human evaluation}
In addition to automated evaluation, we also incorporate human evaluation. Human evaluation not only verifies the reliability of automated evaluation but also addresses complex factors such as contextual information and implicit logic, which are difficult for automated evaluation to capture. The evaluation criteria used for human evaluation are the same as that for automated evaluation. Considering the costs, we employ human evaluation only for content quality evaluation. The details of the human evaluation are presented in the Appendix. \ref{subsec:Details of human evaluation}

\subsection{Experiment Settings}

\paragraph{Implementation} During the retrieval stage and the evaluation stage, we selected bge-base-en-v1.5 \cite{bge_embedding} as the embedding model. Throughout the entire process, we use GPT4o as our LLM agent.\footnote{Specifically, we use gpt-4o-2024-08-06}.
\paragraph{Baselines} We employ the following methods as our baselines:

\textbf{Human}. Human-written surveys collected from arxiv.org. Detailed information can be found at www.surveyx.cn.

\textbf{Navie RAG}: We used the same references as \surveyx{} to provide abstracts for the LLM to guide the generation of the survey. The prompt used by Naive RAG is presented in the Appendix \ref{subsec:Naive RAG Prompt}.

\textbf{AutoSurvey}: An automated survey generation system proposed by \cite{AutoSurvey_24_NIPS_Westlake}. It divides the survey generation process into four stages: Initial Retrieval and Outline Generation, Subsection Drafting, Integration and Refinement, and Rigorous Evaluation and Iteration. In our experiments, we employed its 64k version.

\paragraph{Test Cases} We adopted the 20 topics mentioned by \cite{AutoSurvey_24_NIPS_Westlake} to generate the corresponding surveys for comparison. Details of these topics can be found in the Table \ref{table:survey topics}.

\paragraph{Ablation} To evaluate the impact of different modules or methods on the performance of \surveyx{}, we designed the following ablation setting:

(1) For the retrieval algorithms (component) within the retrieval module of the references acquisition stage, we directly use the initial keywords for retrieval, bypassing the Keyword Expansion retrieval algorithm.

(2) For the AttributeTree method in the references pre-processing stage, we replace the originally generated attribute trees with the full text of the references.

(3) For the outline optimization method in the outline generation stage, we design prompts to let the LLM generate the entire outline in one step.

(4) For the RAG-based rewriting module in the post refinment stage, we eliminate the entire RAG-based rewriting module.

\begin{table*}[!hb]
\caption{Ablation study results for SurveyX with different components removed. Data with significant declines are indicated by underlines.}

\begin{tabular}{cccccccc|ccc}
\hline
Ablation Object        &           & Coverage      & Structure     & Relevance     & Synthesis     & \begin{tabular}[c]{@{}c@{}}Critical\\ Analysis\end{tabular} & Avg            & Recall         & Precision      & F1             \\ \hline
Retrieval Algorithm     &           & {\ul 4.74}    & 4.88          & {\ul 4.79}    & 3.98          & 4.02                                                        & 4.48           & 78.88          & 73.34          & 76.01          \\
AttributeTree Method     &           & 4.84          & {\ul 4.08}    & 4.89          & {\ul 3.88}    & {\ul 3.93}                                                  & 4.32           & {\ul 60.09}    & {\ul 56.49}    & {\ul 58.23}    \\
Outline Optimization Method& \textbf{} & 4.90          & {\ul 3.80}    & 4.91          & 3.98          & 4.02                                                        & 4.32           & 85.1           & 77.13          & 80.92          \\
RAG-based Rewriting Module       &           & 4.92          & 4.89          & 4.93          & 4.00          & 4.00                                                        & 4.55           & {\ul 55.37}    & {\ul 54.95}    & {\ul 55.16}    \\
\textbf{No Ablation} & \textbf{} & \textbf{4.95} & \textbf{4.91} & \textbf{4.94} & \textbf{4.10} & \textbf{4.05}                                               & \textbf{4.590} & \textbf{85.23} & \textbf{78.12} & \textbf{81.52} \\ \hline
\end{tabular}
\label{tab:ablation}

\end{table*}



\subsection{Experiment Results and Analysis}
\begin{table}[h!]
\caption{Evaluation results of reference relevance.}

\resizebox{0.4\textwidth}{!}{
\begin{tabular}{ccccc}
\hline
\multicolumn{1}{l}{Model} & \multicolumn{1}{l}{} & \multicolumn{1}{l}{IoU} & \multicolumn{1}{l}\text{Relevance}\textsubscript{\textit{semantic}} & \multicolumn{1}{l}\text{Relevance}\textsubscript{\textit{LLM}} \\ \hline
Human                     &                      & 1                       & 0.4455                                      & 0.9485                                \\
\surveyx{}                   &                      & 0.55                    & 0.4226                                      & 0.7689                                \\ \hline
\end{tabular}
}
\label{tab:reference}
\end{table}
\paragraph{Main results} 
The results for content quality evaluation, citation quality evaluation, and reference relevance evaluation are shown in Table \ref{tab:content} and Table \ref{tab:reference}. Key findings are as follows: 

(1) The experimental results for content quality evaluation indicate that \surveyx{} performs exceptionally well across all metrics, particularly in Coverage (4.95), Structure (4.91), and Relevance (4.94), closely approaching the performance of human experts. This demonstrates that the surveys generated by \surveyx{} not only comprehensively cover both the core and peripheral content of the given topic but also ensure logical and coherent outlines while maintaining high relevance. Compared to naive RAG and AutoSurvey, \surveyx{} shows a clear advantage across all metrics, with significant improvements in Structure (4.91) and Critical Analysis (4.05). This indicates that the surveys generated by \surveyx{} have clearer and more coherent outlines and greater depth in content. Additionally, \surveyx{} leads existing automated survey generation systems with an average score of 4.590, proving its outstanding performance in survey generation.

(2) The results of the citation quality experiments show that \surveyx{} outperforms existing automated survey generation systems and closely approaches the performance of human experts in Citation Recall (85.23), Citation Precision (78.12), and F1 Score (81.52). Notably, in the Precision metric (78.12), \surveyx{} even slightly surpasses human experts. This indicates that \surveyx{} can significantly enhance the comprehensiveness and accuracy of citations in its generated surveys, ensuring that all statements are well-supported by literature while effectively reducing irrelevant citations, thereby increasing the reliability and credibility of the generated surveys. These experimental results also indirectly demonstrate that the RAG-based rewriting component used in the post refinment stage of \surveyx{} effectively improves the citation quality of the generated surveys.

(3) The results of the reference relevance experiments indicate that \surveyx{} approaches the performance of human experts in the \text{Relevance}\textsubscript{\textit{semantic}} metric. However, there is a certain gap compared to human experts in the IoU and \text{Relevance}\textsubscript{\textit{LLM}} metrics. Nonetheless, as the first automated survey generation system in this field with comprehensive online reference retrieval capabilities, \surveyx{} demonstrates its potential and application prospects, laying the foundation for future research.

\begin{figure}[h]
  \centering
  \includegraphics[width=\columnwidth]{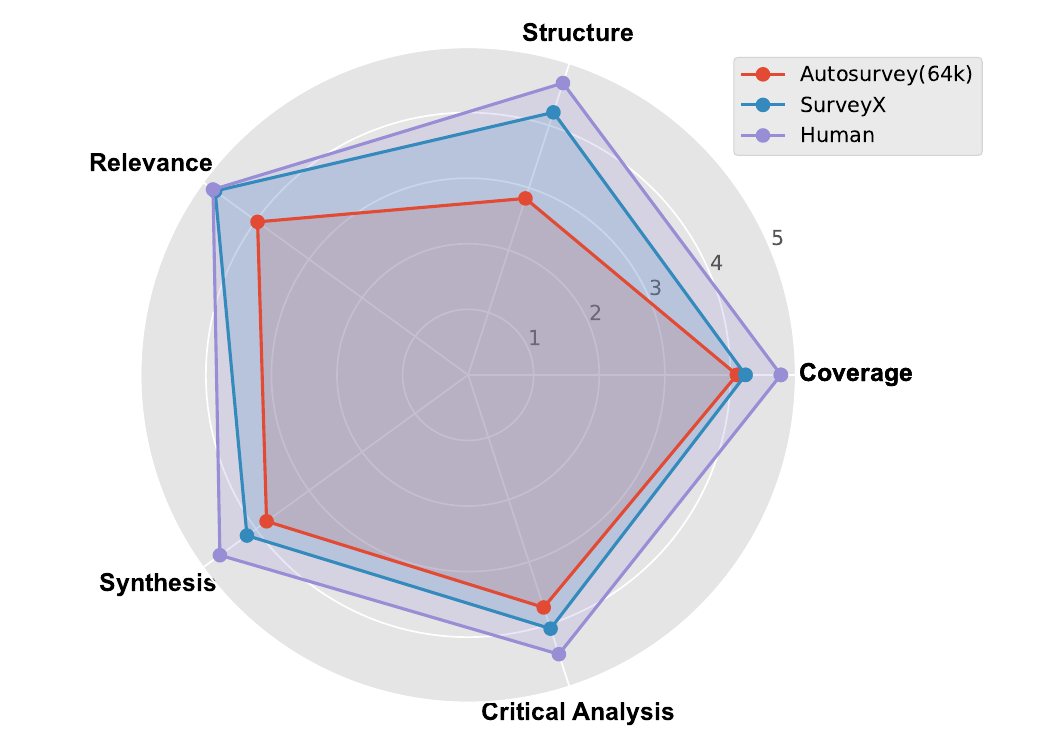}
  \caption{Human evaluation results.}
  \label{fig:human_eval}
\end{figure}

\paragraph{Human evaluation results}
The results of the human evaluation are shown in Figure \ref{fig:human_eval}. The results show that \surveyx{} outperforms AutoSurvey across all metrics and is closer to the performance of human experts. This aligns with the results of the automated evaluation, which to some extent supports the validity of our automated evaluation method. Additionally, compared to the automated evaluation results, the scores from human evaluation are generally lower, especially in the Structure metric. This reflects the higher standards of human evaluators for the survey in areas such as structural coherence, logical consistency, content depth, etc.

\paragraph{Ablation results}
The results of the ablation experiments are shown in Table \ref{tab:ablation}. The experimental results indicate that each module of \surveyx{} plays a crucial role. Key findings are as follows: 

(1) After ablating the retrieval algorithm, the Coverage (4.95→4.74) and Relevance (4.94→4.79) metrics showed the most significant decreases. This indicates that the improvement in reference quality brought by the retrieval algorithm is crucial for ensuring comprehensive content coverage and high topic relevance in the generated surveys.

(2) After ablating the AttributeTree method, the Structure (4.91→\newline4.08), Synthesis (4.10→3.88), and Critical Analysis (4.05→3.93) metrics showed significant declines. Besides, the Recall (85.23→60.09), Precision (78.12→56.49), and F1 score (81.52→58.23) metrics also decreased noticeably. This indicates that converting the content of the reference into an attribute tree enables the LLM to understand the core information better, thereby improving its ability to organize outlines, integrate information from different references, conduct in-depth analysis, and add references more accurately.

(3) After ablating the outline optimization method, the Structure metric (4.91→3.80) showed the most significant decline. This indicates that this method effectively enhances the logical coherence and structural clarity of the generated survey outline.

(4) After ablating the RAG-based rewriting module, the Recall (85.23→55.37), Precision (78.12→54.95), and F1 score (81.52→55.16) metrics declined significantly. This indicates that the RAG-based rewriting module can considerably enhance the citation quality of the generated survey by more accurately adding relevant citations and removing irrelevant ones.

\section{Conclusion}
In this paper, we propose \surveyx{}, an innovative automated survey generation system. \surveyx{} effectively addresses issues in LLM-based automated survey generation, such as context window limitations and internal knowledge constraints. Moreover, \surveyx{} also overcomes shortcomings of existing automatic generation systems, including a lack of diversity in survey expression, inadequate reference pre-processing methods, and limitations in retrieval approaches. Experimental results demonstrate that \surveyx{} significantly outperforms existing systems in both content and citation quality, approaching the level of human experts. This achievement indicates that \surveyx{} can serve as a reliable and efficient tool for automated survey generation, providing valuable assistance to researchers.

In the future, we consider improving our system in the following areas:

\begin{itemize}
\item Optimizing the retrieval algorithm to achieve retrieval performance comparable to human levels.
\item Expanding the methods for generating figures and tables.
\item Enhancing the organization of surveys by further refining the composing approach based on the attribute tree.
\end{itemize}

\bibliographystyle{ACM-Reference-Format}
\bibliography{reference}

\appendix
\section{Evaluation details}
\label{sec:Evaluation details}

\subsection{Survey Topics}
\label{subsec:Survey Topics}

Table \ref{table:survey topics} presents the 20 survey topics used in the evaluation experiment, along with the titles of the corresponding survey written by human experts.
\begin{table*}[h!]
    \caption{Survey Topics}
    \resizebox{\textwidth}{!}{ 
    \begin{tabular}{ll} 
        \toprule
        \textbf{Topic} & \textbf{Survey Title}  \\ 
        \midrule
         In-context Learning& A survey for in-context learning  \\
         LLMs for Recommendation& A Survey on Large Language Models for Recommendation  \\
         LLM-Generated Texts Detection& A Survey of Detecting LLM-Generated Texts  \\
         Explainability for LLMs& Explainability for Large Language Models  \\
         Evaluation of LLMs& A Survey on Evaluation of Large Language Models \\
         LLMs-based Agents& A Survey on Large Language Model based Autonomous Agents  \\
         LLMs in Medicine& A Survey of Large Language Models in Medicine  \\
         Domain Specialization of LLMs& Domain Specialization as the Key to Make Large Language Models Disruptive  \\
         Challenges of LLMs in Education& Practical and Ethical Challenges of Large Language Models in Education \\
         Alignment of LLMs& Aligning Large Language Models with Human  \\
         ChatGPT& A Survey on ChatGPT and Beyond  \\
         Instruction Tuning for LLMs& Instruction Tuning for Large Language Models  \\
         LLMs for Information Retrieval& Large Language Models for Information Retrieval \\
         Safety in LLMs& Towards Safer Generative Language Models: Safety Risks, Evaluations, and Improvements  \\
         Chain of Thought& A Survey of Chain of Thought Reasoning  \\
         Hallucination in LLMs& A Survey on Hallucination in Large Language Models  \\
         Bias and Fairness in LLMs& Bias and Fairness in Large Language Models  \\
         Large Multi-Modal Language Models& Large-scale Multi-Modal Pre-trained Models  \\
         Acceleration for LLMs& A Survey on Model Compression and Acceleration for Pretrained Language Models \\
         LLMs for Software Engineering& Large Language Models for Software Engineering  \\
        \bottomrule
    \end{tabular}
    }
    \label{table:survey topics}
\end{table*}

\subsection{Details of human evaluation}
\label{subsec:Details of human evaluation}
For the human evaluation, six PhD students with experience in writing LLM-related surveys were invited. The platform used for evaluation is Label Studio\footnote{\url{https://labelstud.io/}}. The final score was derived by averaging the scores given by all evaluators.
\subsection{Evaluation Prompt}
\label{subsec:Evaluation Prompt}
\begin{figure*}[h]
  \centering
    \includegraphics[width=\textwidth]{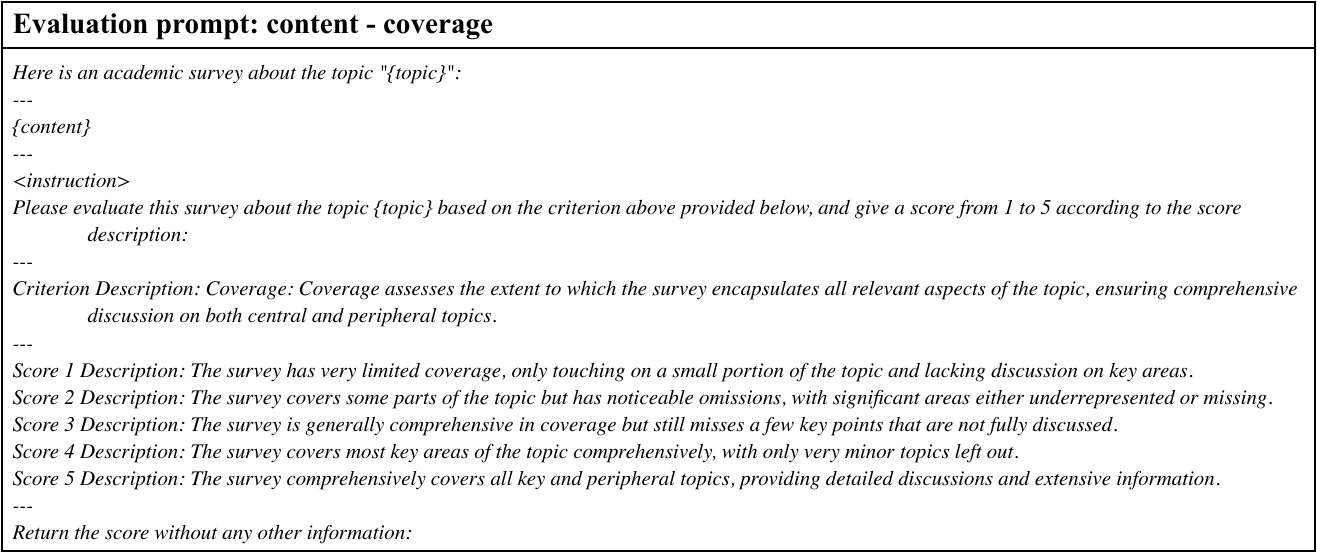}
  \caption{Content coverage prompt for evaluation.}
  \label{fig:eval_content_coverage}
\end{figure*}

\begin{figure*}[h]
  \centering
    \includegraphics[width=\textwidth]{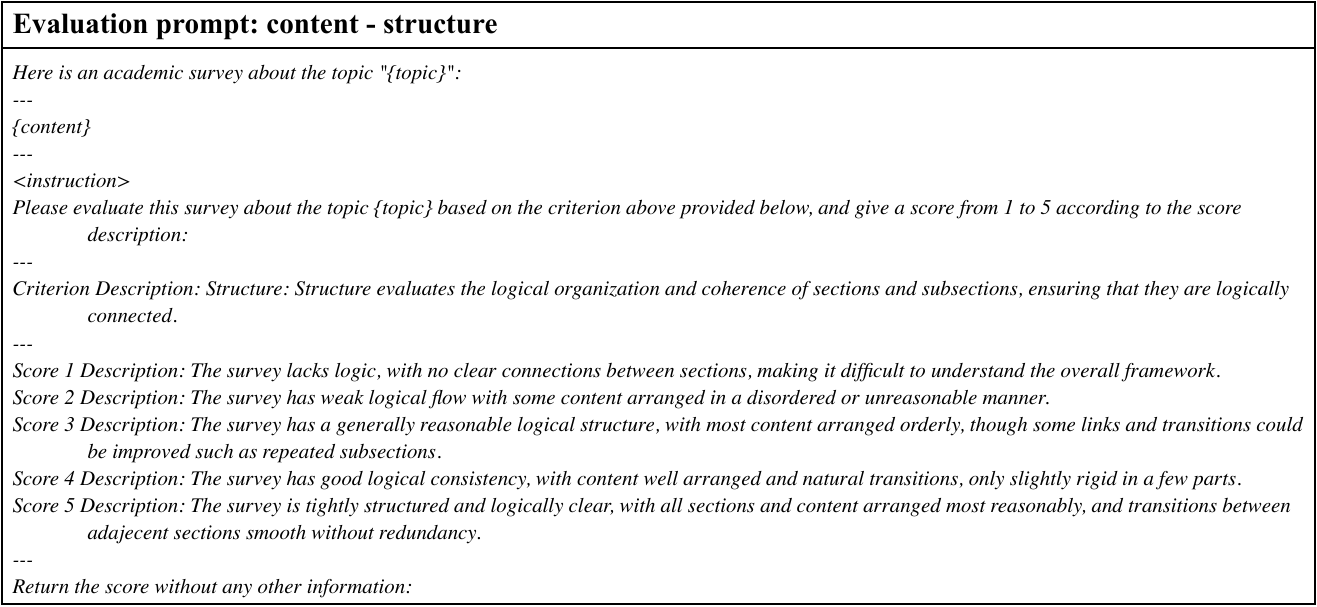}
  \caption{Content structure prompt for evaluation.}
  \label{fig:eval_content_structure}
\end{figure*}

\begin{figure*}[h]
  \centering
    \includegraphics[width=\textwidth]{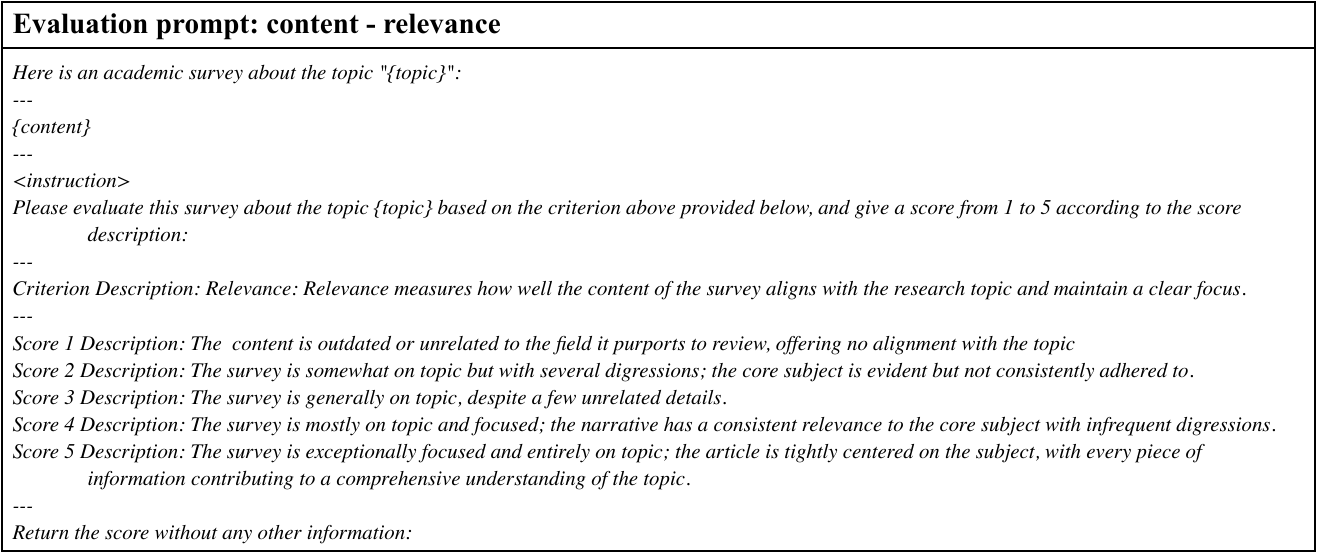}
  \caption{Content relevance prompt for evaluation.}
  \label{fig:eval_content_relevance}
\end{figure*}

\begin{figure*}[h]
  \centering
    \includegraphics[width=\textwidth]{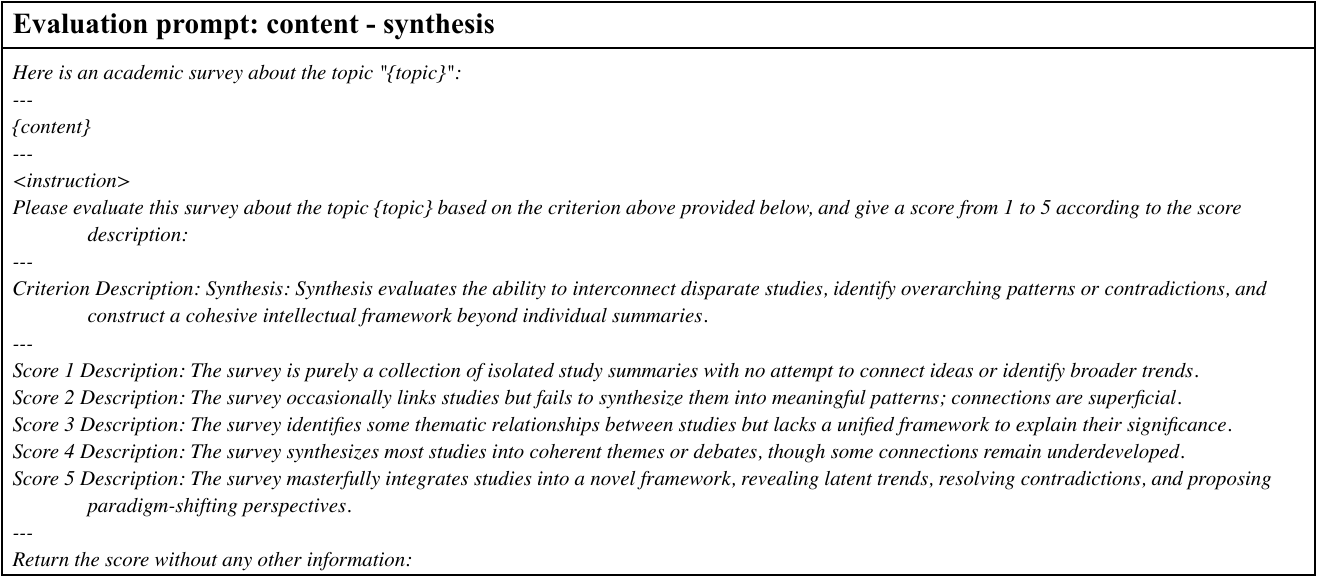}
  \caption{Content synthesis prompt for evaluation.}
  \label{fig:eval_content_synthesis}
\end{figure*}

\begin{figure*}[h]
  \centering
    \includegraphics[width=\textwidth]{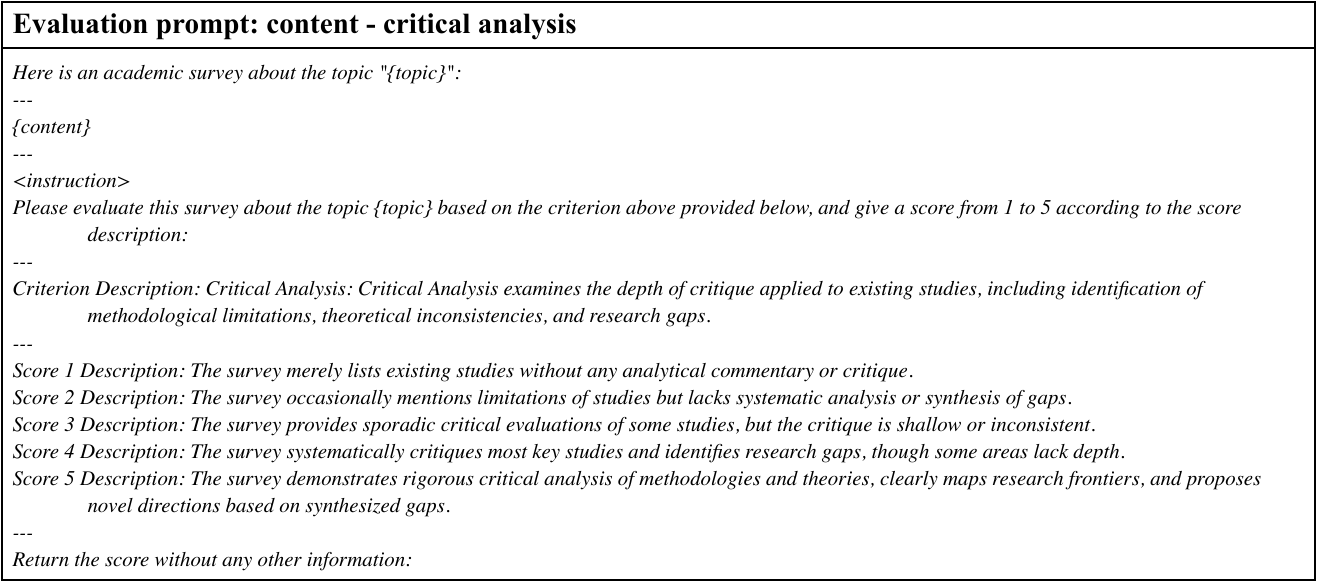}
  \caption{Content critical analysis prompt for evaluation.}
  \label{fig:eval_content_ca}
\end{figure*}

\begin{figure*}[h]
  \centering
    \includegraphics[width=\textwidth]{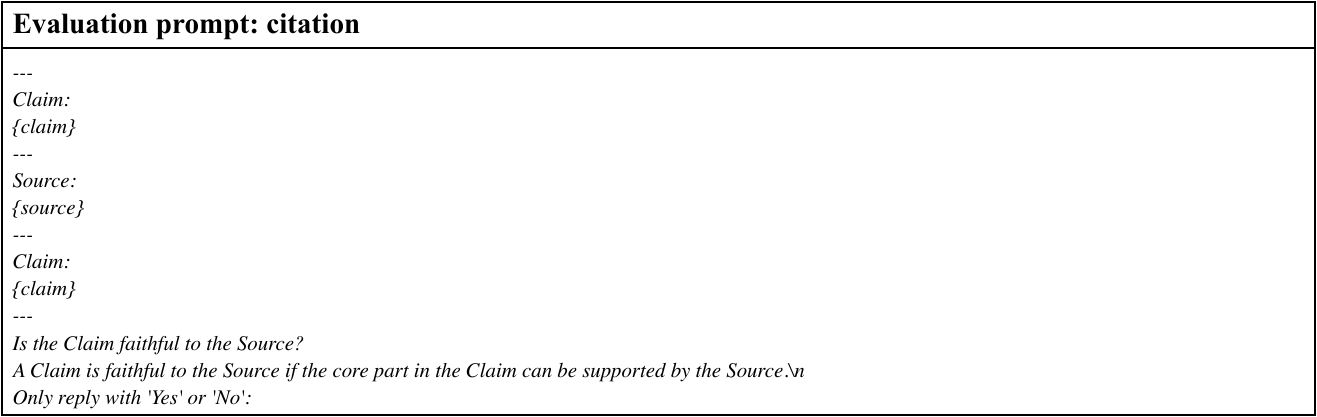}
  \caption{Citation prompt for evaluation.}
  \label{fig:eval_citation}
\end{figure*}

\begin{figure*}[h]
  \centering
    \includegraphics[width=\textwidth]{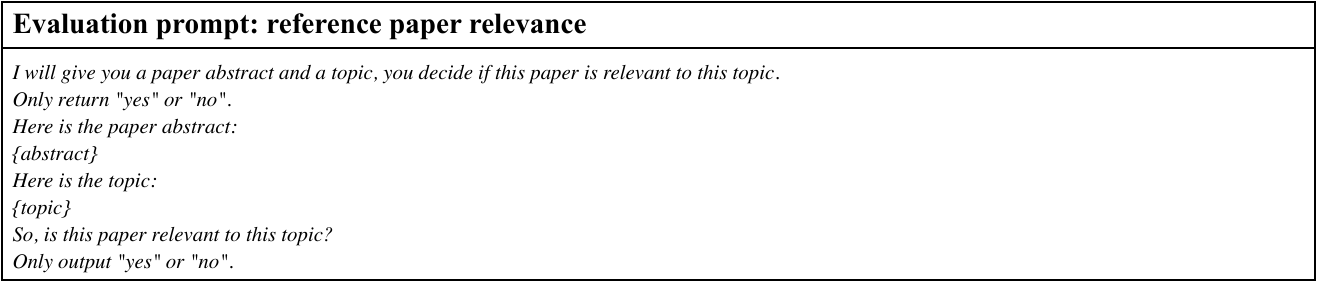}
  \caption{Judge relevance for reference paper.}
  \label{fig:eval_ref}
\end{figure*}

Prompts used for content evaluation are presented in figure \ref{fig:eval_content_coverage}, \ref{fig:eval_content_structure}, \ref{fig:eval_content_relevance},
\ref{fig:eval_content_synthesis},
\ref{fig:eval_content_ca}.
Prompts used for citation quality are presented in figure \ref{fig:eval_citation}.
Prompts used for reference papers are presented in figure \ref{fig:eval_ref}.

\subsection{Naive RAG Prompt}
\label{subsec:Naive RAG Prompt}
\begin{figure*}[h]
  \centering
    \includegraphics[width=\textwidth]{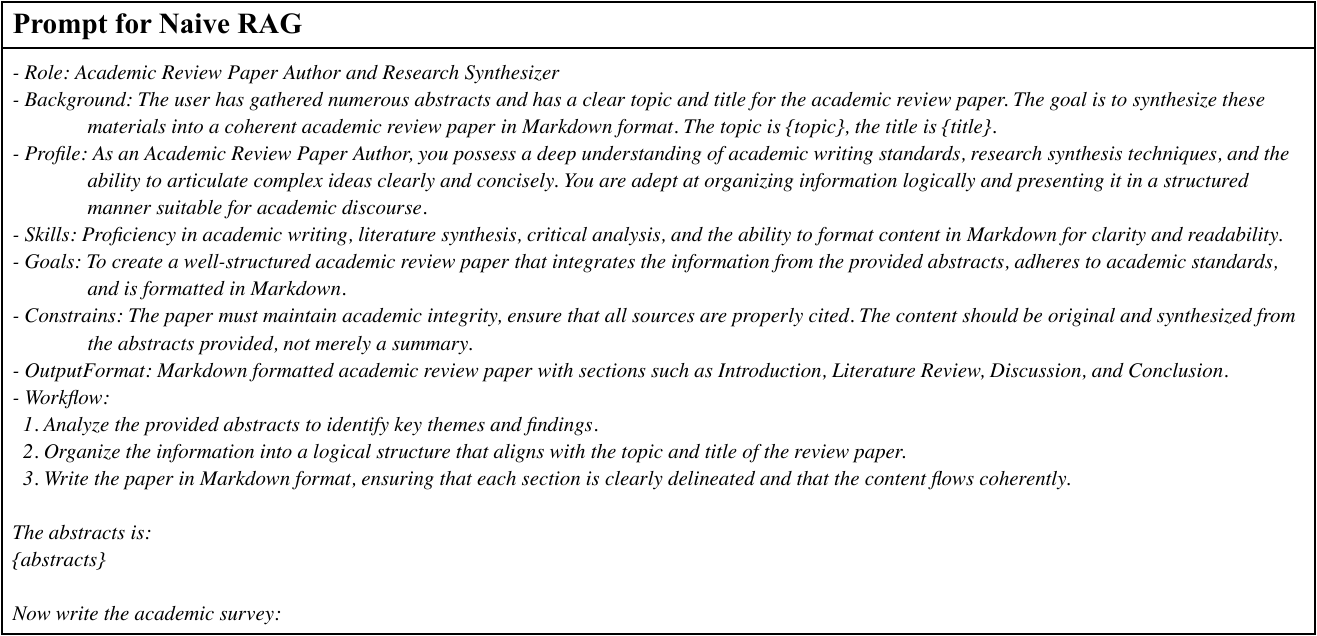}
  \caption{Prompts for Naive RAG.}
  \label{fig:naive_rag_prompt}
\end{figure*}

Prompts used for Naive RAG method are presented in figure \ref{fig:naive_rag_prompt}
\section{Attribute Tree Templates}
\label{sec:Attribute Tree Templates}
\begin{figure*}[h]
  \centering
  \includegraphics[width=\textwidth]{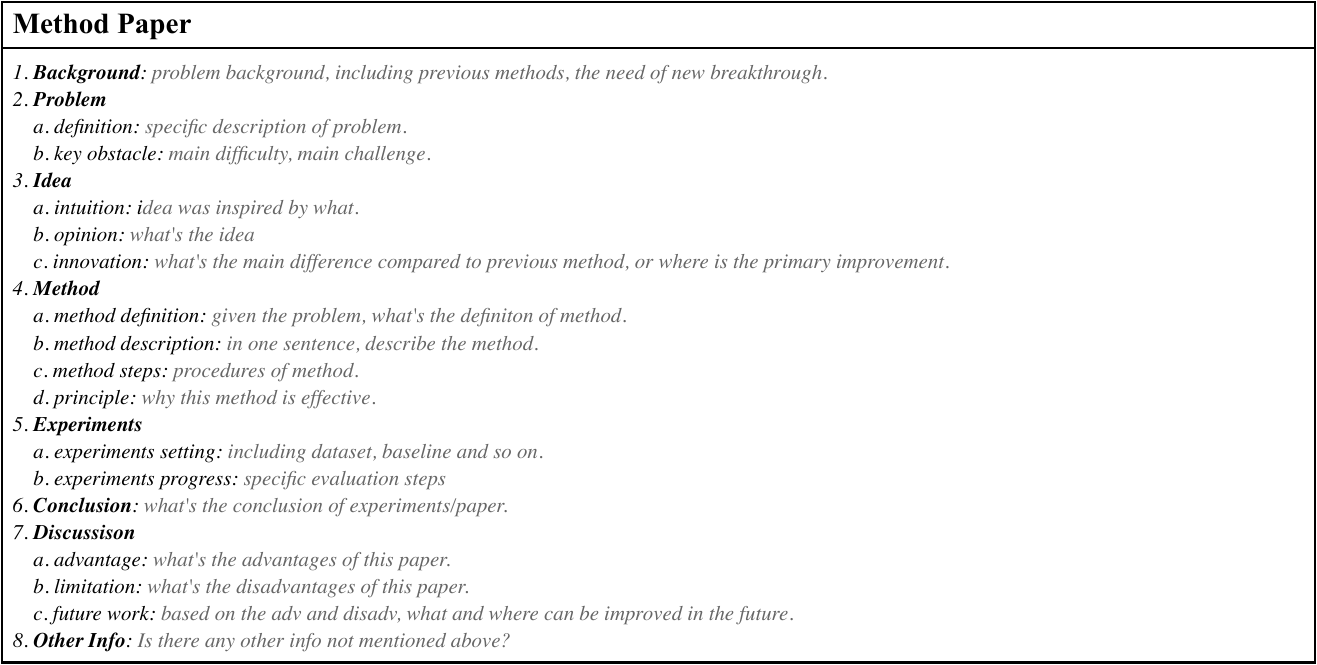}
  \caption{Method paper attribute tree.}
  \label{fig:attribute_tree_method}
\end{figure*}

\begin{figure*}[h]
  \centering
    \includegraphics[width=\textwidth]{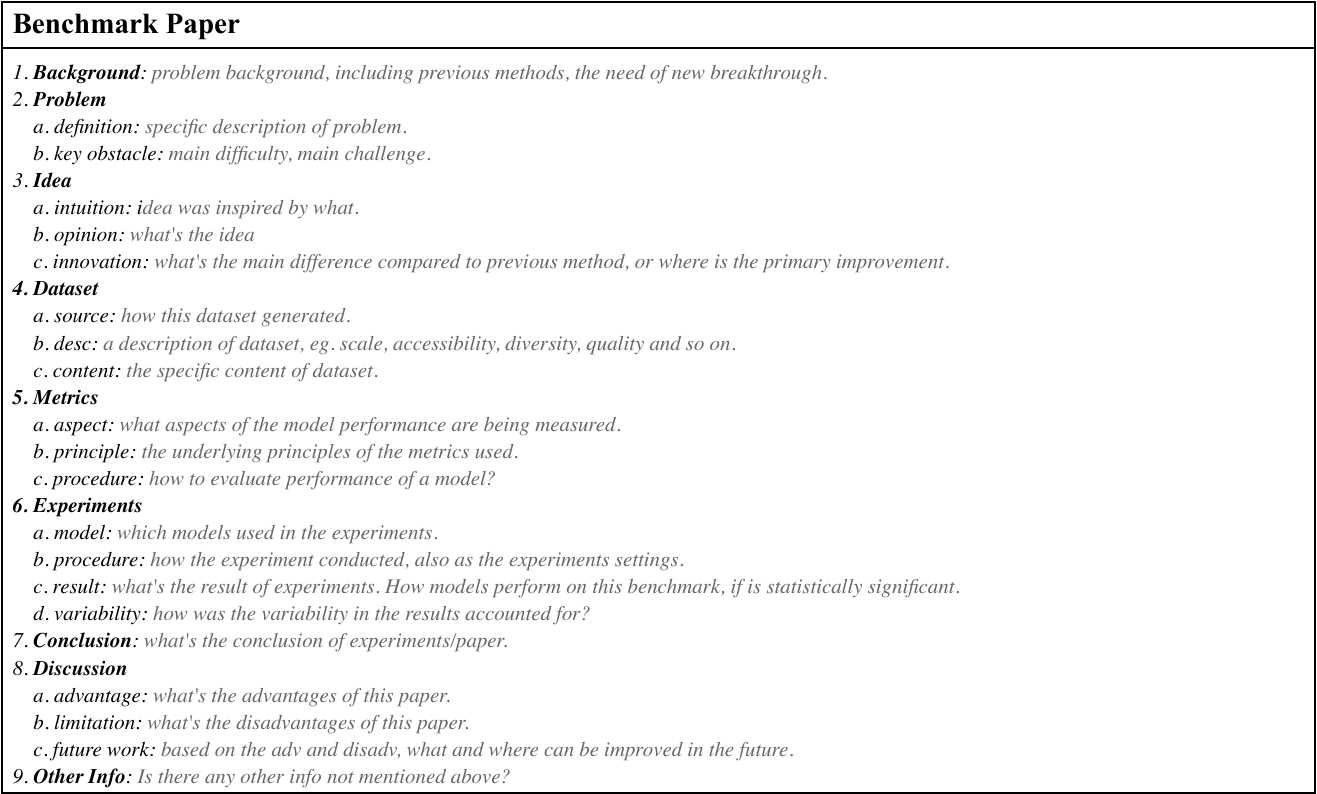}
  \caption{Benchmark paper attribute tree.}
  \label{fig:attribute_tree_benchmark}
\end{figure*}

\begin{figure*}[h]
  \centering
    \includegraphics[width=\textwidth]{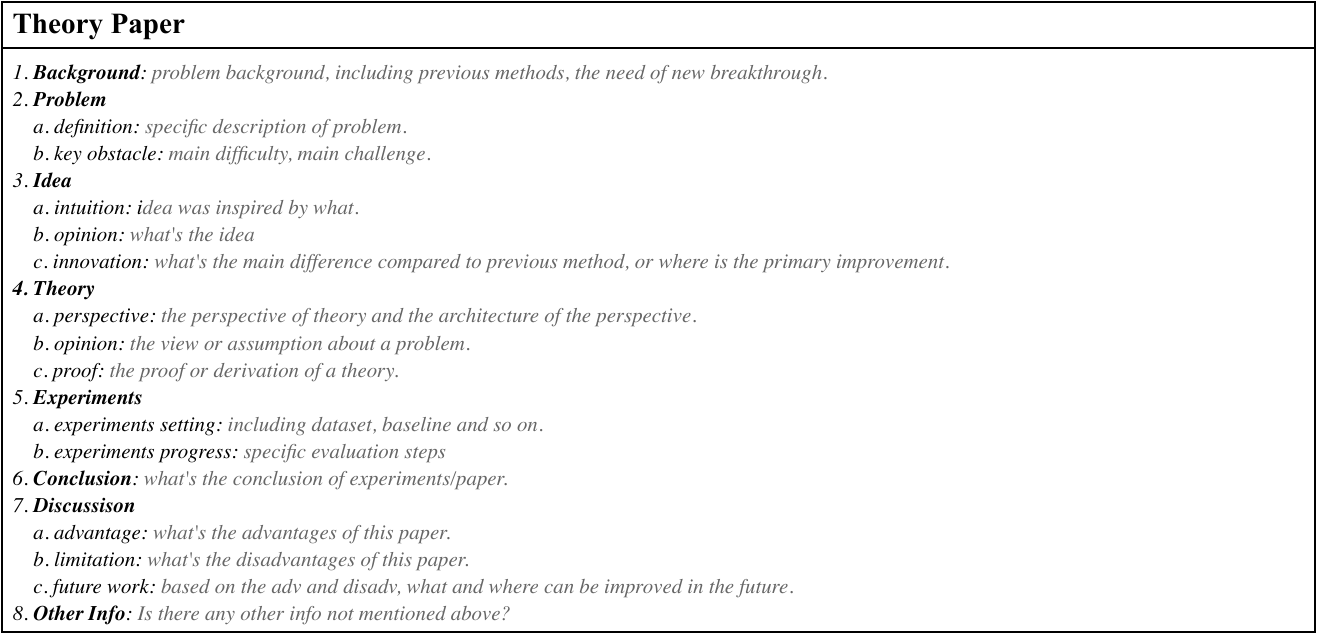}
  \caption{Theory paper attribute tree.}
  \label{fig:attribute_tree_theory}
\end{figure*}

\begin{figure*}[h]
  \centering
    \includegraphics[width=\textwidth]{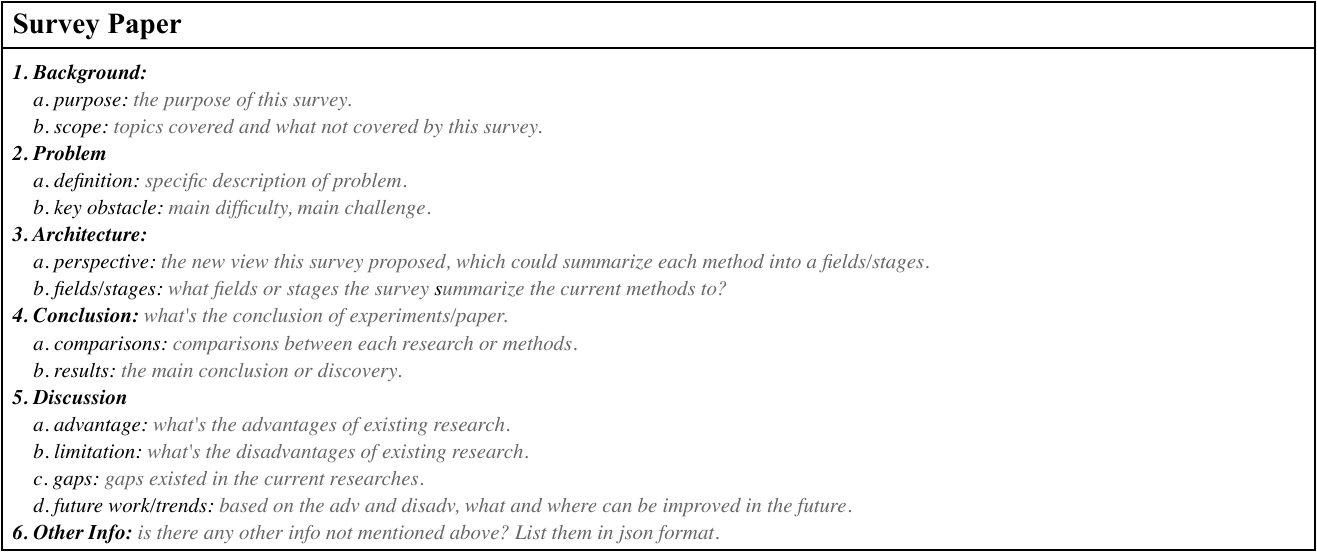}
  \caption{Survey paper attribute tree.}
  \label{fig:attribute_tree_survey}
\end{figure*}

Attribute tree templates are shown in figure \ref{fig:attribute_tree_method}, \ref{fig:attribute_tree_benchmark}, \ref{fig:attribute_tree_theory}, 
\ref{fig:attribute_tree_survey}.

\end{document}